\documentclass[10pt,twocolumn,letterpaper]{article}

\usepackage[pagenumbers]{cvpr} 

\usepackage{graphicx}
\usepackage{amsmath}
\usepackage{xcolor}
\usepackage{colortbl}
\usepackage{amssymb}
\usepackage{booktabs}

\usepackage[pagebackref,breaklinks,colorlinks]{hyperref}

\usepackage[capitalize]{cleveref}
\crefname{section}{Sec.}{Secs.}
\Crefname{section}{Section}{Sections}
\Crefname{table}{Table}{Tables}
\crefname{table}{Tab.}{Tabs.}

\usepackage{bbding}
\usepackage{xcolor}
\definecolor{green}{RGB}{0,255,0}
\definecolor{red}{RGB}{255,0,0}
\usepackage{multirow}
\usepackage{stfloats}
\usepackage{bm}
\def\cvprPaperID{8383} 
\def\confName{CVPR}
\def\confYear{2023}

\begin{document}

\title{Meta Architecture for Point Cloud Analysis}



\author{
    Haojia Lin$^{1}$,
    Xiawu Zheng$^{2}$,
    Lijiang Li$^{1}$,
    Fei Chao$^{1}$,
    Shanshan Wang$^{3}$,\\
    Yan Wang$^{4}$,
    Yonghong Tian$^{2, 5}$,
    Rongrong Ji$^{1, 2}$\thanks{Corresponding author: rrji@xmu.edu.cn}, \\
    $^1$Media Analytics and Computing Lab, Department of Artificial Intelligence, School of Informatics, \\Xiamen University. 
    $^2$Peng Cheng Laboratory. 
    $^3$Chinese Academy of Sciences. 
    $^4$Samsara Inc.\\
    $^5$National Engineering Research Center for Visual Technology, Peking University. \\
    {\tt\small \{linhj, lilijiang\}@stu.xmu.edu.cn, zhengxw01@pcl.ac.cn, \{rrji, feichao\}@xmu.edu.cn}\\
    {\tt\small ss.wang@siat.ac.cn, grapeot@outlook, yhtian@pku.edu.cn}
}
\maketitle

\begin{abstract}
Recent advances in 3D point cloud analysis bring a diverse set of network architectures to the field.
However, the lack of a unified framework to interpret those networks makes any systematic comparison, contrast, or analysis challenging, and practically limits healthy development of the field.
In this paper, we take the initiative to explore and propose a unified framework called PointMeta, to which the popular 3D point cloud analysis approaches could fit. 
This brings three benefits.
First, it allows us to compare different approaches in a fair manner, and use quick experiments to verify any empirical observations or assumptions summarized from the comparison.
Second, the big picture brought by PointMeta enables us to think across different components, and revisit common beliefs and key design decisions made by the popular approaches.
Third, based on the learnings from the previous two analyses, by doing simple tweaks on the existing approaches, we are able to
derive a basic building block, termed PointMetaBase.
It shows very strong performance in efficiency and effectiveness through extensive experiments on challenging benchmarks, and thus verifies the necessity and benefits of high-level interpretation, contrast, and comparison like PointMeta.
In particular, PointMetaBase surpasses the previous state-of-the-art method by 0.7\%/1.4/\%2.1\% mIoU with only 2\%/11\%/13\% of the computation cost on the S3DIS datasets. 
The code and models are available at  \url{https://github.com/linhaojia13/PointMetaBase}.
\end{abstract}


\section{Introduction}
In the past two decades, the popularity of 3D data acquisition technology has led to great interest in point cloud analysis.
Unlike images, point clouds are inherently sparse, unordered, and irregular. 
These characteristics make it challenging for the powerful convolutional neural networks (CNN) \cite{resnet,vgg} to extract useful information from point clouds.
Early studies attempt to convert the point cloud into grids by either voxelization or 2D projections, such that the standard CNN can be applied, but the applications are limited by the extra computation and information loss.

\begin{figure}[t!]
  \centering 
  \includegraphics[width=1.0\linewidth]{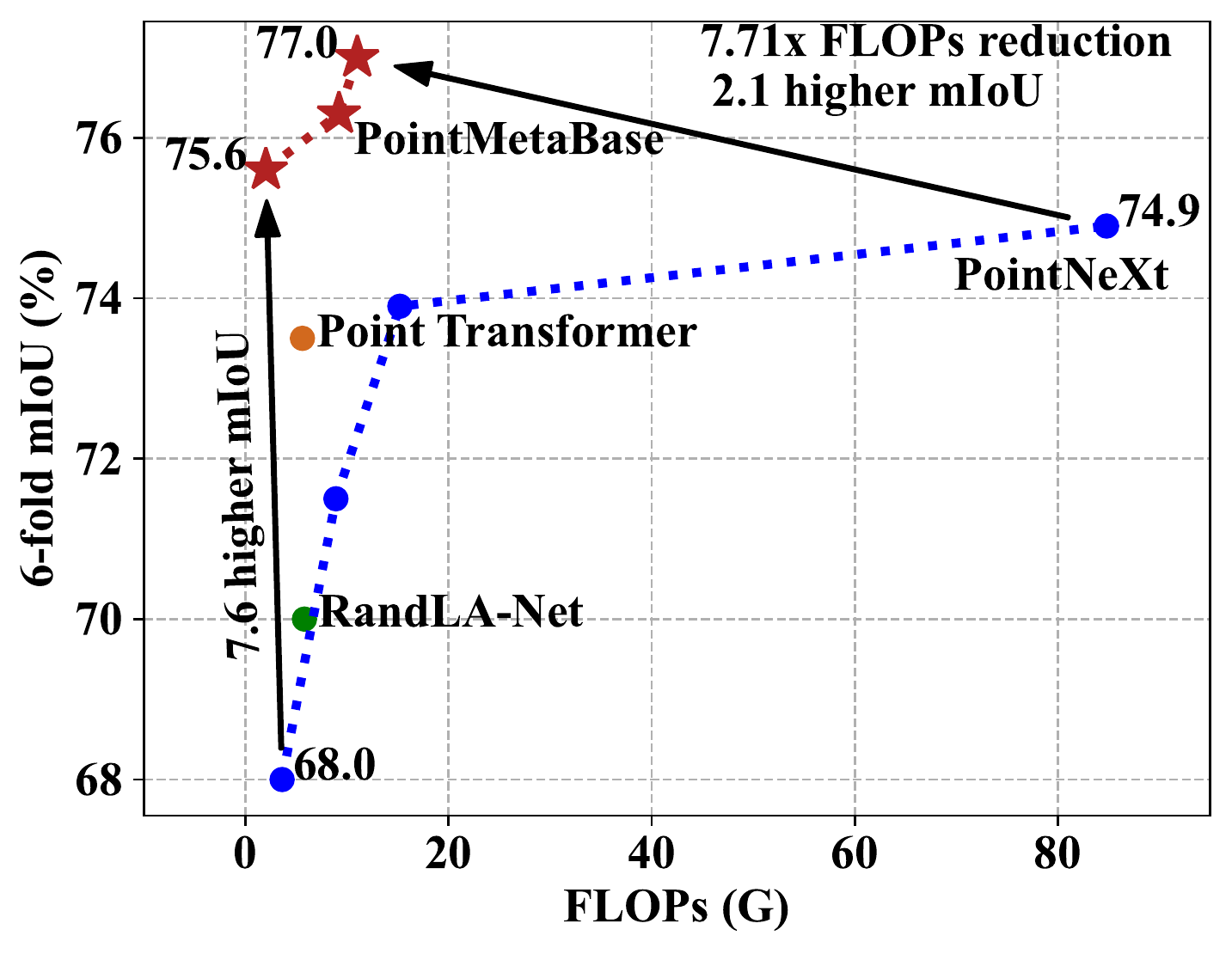}
  \vspace{-2.2em}
  \caption{Segmentation performance of PointMetaBase on S3DIS \cite{s3dis}. 
  PointMetaBase surpasses the state-of-the-art method PointNeXt \cite{pointnext}
  significantly with a large FLOPs reduction. }
  \label{flops_6foldmiou} 
\end{figure}

With the introduction of permutation equivariance and invariance by PointNet \cite{pointnet} and PointNet++ \cite{pointnet2}, 
it is possible to use CNNs to process point clouds in their unstructured format. 
Following PointNet++ \cite{pointnet2}, many point-based networks have been introduced, 
most of which focus on the development of complex building blocks to extract local features, 
such as the $\mathcal{X}-\text{Conv}$ convolution in PointCNN \cite{kpconv} 
and the self-attention layer in Point Transformer \cite{pointtransformer}. 
Although brought significant performance gains, 
these models are very complicated.
On one hand, this makes the approaches computational demanding and thus limited in applications.
On the other hand, it also gives extra challenges in systematic comparison and analysis among the models, and thus impacts efficient development of the field, which ideally would benefit from theoretic and empirical guidance.

The community is well aware of this issue, and research efforts have been put on looking for a unified perspective to compare these methods and identify the most crucial implementation details. 
PosPool \cite{pospool} and PointNeXt \cite{pointnext} take a step toward this goal. 
Adopting a deep residual architecture as the base network, 
PosPool \cite{pospool} evaluates the representative building blocks \cite{pointnet2,pointcnn,kpconv,dpc} 
and finds that they perform similarly well. 
PointNeXt \cite{pointnext} revisits the improved training and scaling strategies adopted 
by previous SotA methods \cite{deepgcn,kpconv,pointtransformer,assanet}
and tweaks the early model PointNet++ \cite{pointnet2} with the learnings, 
which achieves state-of-the-art performance. 
Both of them give insights and inspirations to the community 
via dedicated experiment designs and extensive empirical analysis. 
However, neither of them provides a framework that is general enough to fit a broad range of point cloud analysis approaches, such that large scale comparison and contrast could be done.

In this paper, we argue that if viewed from the perspective of basic building blocks, the majority of existing approaches could be fit into  a single meta architecture (Sec. \ref{sec_pointmeta}).
We call it PointMeta.
PointMeta abstracts the computation pipeline of the building blocks for point cloud analysis 
into four meta functions: a neighbor update function, a neighbor aggregation function, 
a point update function and a position embedding function (implicit or explicit). 
As will be discussed in more details in the next sections, this framework brings three benefits, which are also our core technical contributions:

\begin{itemize}
    \item In the \textbf{dimension of models}, it allows us to compare and contrast different models in a fair manner. So it becomes practical to observe and summarize learnings and assumptions, whose correctness could be further verified through experiments with variables controlled under the same framework. For example, among the systematic analysis on all the components across popular models, we found for the positional embedding function, explicit positional embedding is empirically the best choice (Sec. \ref{sec_pe}).
    \item In the \textbf{dimension of components}, it allows us to have a higher level view across components, and thus to revisit the common beliefs and design decisions of the existing approaches. For example, despite the common perception, we find that the neighbor aggregation blocks may collaborate or compete with the learned neighbor features (Sec. \ref{sec_na}), and thus should be designed carefully.
    \item Based on the learnings from the previous two dimensions, we are then able to do simple tweaks on the building blocks to \textbf{apply the best practices} (Sec. \ref{sec_practice}). The result building block, PointMetaBase, achieves very strong performance, surpassing the previous state-of-the-art method~\cite{pointnext} by 0.7\%/1.4/\%2.1\% mIoU with only 2\%/11\%/13\% of the computation cost on S3DIS~\cite{s3dis} (Fig. \ref{flops_6foldmiou}), and can act as a baseline for further research.
\end{itemize}


\section{Related Work}
\label{sec_related}

\textbf{Neural Networks for Point Cloud Analysis.} 
The unstructured data format of point cloud makes it difficult to apply CNN directly. 
To address this challenge,
PointNet \cite{pointnet} introduces permutation equivariance and invariance 
via point-wise MLP and max pooling, 
which makes it possible to process point cloud directly.    
To better capture locality, PointNet++ \cite{pointnet2} presents a Set Abstraction module 
to aggregate features from the neighbor points. 
Most works after PointNet++ focus on the design of build blocks.
Graph-based methods \cite{dgcnn,gac,deepgcn} utilize graph neural networks to extract local features. 
Pseudo-grid-based methods \cite{pointcnn, tangent, kpconv} transform the local features onto pseudo grids which allow for regular convolutions. 
Adaptive weight-based methods \cite{randla,RSNet} aggregate the neighbor features 
via weights adaptively learned from the locality. 
Recently, Transformer-based block \cite{pointtransformer,stratified} are introduced 
to capture local information through self-attention. 
These proposed building blocks are so complicated that make challenges in systematic comparison and analysis, 
which suggests an urgent need of a common framework 
to unify existing models in this field. 


\begin{figure*}[h!]
  \centering 
  \includegraphics[width=1.0\linewidth]{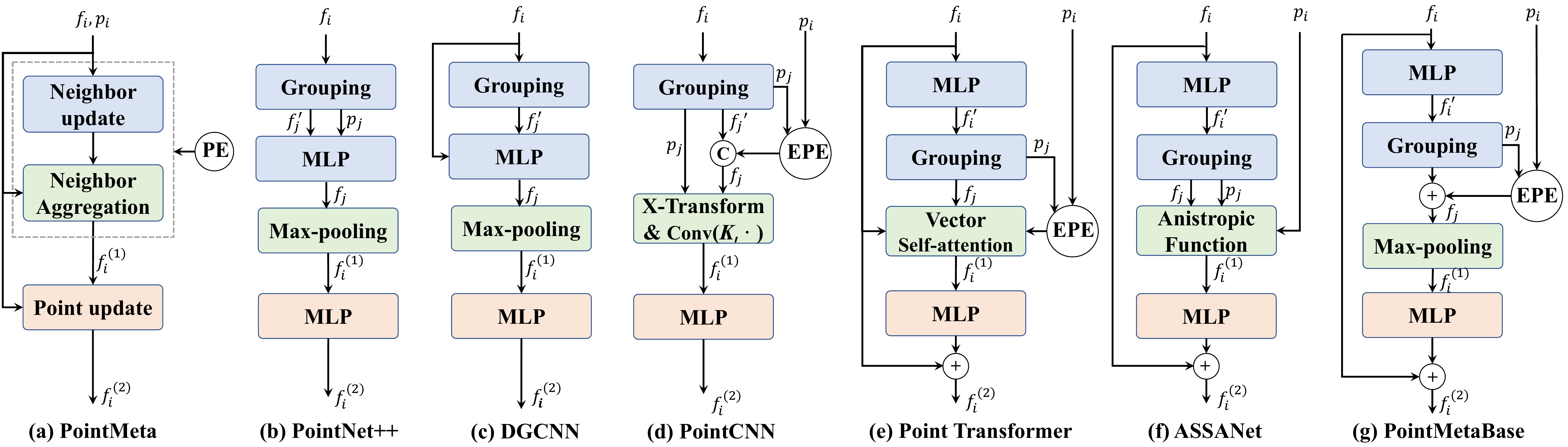}
  \hspace{-1.0em}
  \vspace{-0.6em}
  \caption{PointMeta and its instantiation examples. 
  (a) In PointMeta, the position embedding function is usually combined with 
  the neighbor update function or the aggregation function implicitly or explicitly. 
  (b)$\sim$(f) Representive building blocks  can naturally fit into PointMeta. 
  (g) Applying the summarized best practices, we do simple tweaks on the building blocks 
  and propose PointMetaBase. }
  \label{fig_meta_arch} 
\end{figure*}

\section{PointMeta}
\label{sec_pointmeta}
\subsection{Meta Architecture} \label{sec_meta_arch}
Inspired by the graph network framework 
\cite{RIBGN}, 
we reformulate existing  building blocks for point cloud analysis 
into a single meta architecture, named PointMeta. 
As shown in Fig. \ref{fig_meta_arch} (a), a PointMeta block is composed of four meta functions: 
a neighbor update function $\phi^{n}$, a neighbor aggregation function $\phi^{a}$, 
a point update function $\phi^{p}$ and a position embedding function $\phi^{e}$ (implicit or explicit).
The forward pass of a PointMeta block is as follows: 
\begin{equation} \label{nu}
  \bm{f}_{\mathcal{N}(i)} = \phi^{n}\circ \phi^{e}(f_i, p_i), 
\end{equation}
\begin{equation}
  f_i^{(1)} = \phi^{a}\circ \phi^{e}(f_i, p_i, \bm{f}_{\mathcal{N}(i)}, \bm{p}_{\mathcal{N}(i)})
\end{equation}
\begin{equation}
  f_i^{(2)} = \phi^{p}(f_i^{(1)}, p_i),
\end{equation}
where $f_i\in\mathbb{R}^{d}$ and $p_i\in\mathbb{R}^{3}$ represent the feature and the coordinate of point $i$ respectively, 
and $d$ denotes the feature length. 
Let $\mathcal{N}(i)$ denote the neighbors of point $i$, 
$f_j$ and $p_j$ be the feature and coordinate of each neighbor of point $i$ respectively,   
$\bm{f}_{\mathcal{N}(i)}=\{f_j | j\in \mathcal{N}(i)\}$ represent all neighbor features of point $i$, 
$\bm{p}_{\mathcal{N}(i)}=\{p_j | j\in \mathcal{N}(i)\}$ represent all neighbor coordinates of point $i$, 
and $k$ denote the neighbor number.

\textbf{Neighbor update.} 
The first step is to group and update neighbors for each point. 
Usually, the neighbor features $\bm{f}_{\mathcal{N}(i)}$ 
is produced by K-Nearest query as \cite{dgcnn, pointtransformer}
or ball query as \cite{pointnet2,assanet,pointnext}. 
The update on neighbor features is usually conducted by a multi-layer perceptions (MLP)
to make sure the permutation equivariance of the neighbor update function $\phi^{n}$. 
Specifically, a permutation transformation on $\bm{f}_{\mathcal{N}(i)}$ 
results in the same permutation change on the output of the neighbor update. 

\textbf{Neighbor aggregation.} 
Literally, the neighbor aggregation function is used to 
aggregate the neighbor features for a center point. 
Different from $\phi^{n}$, the aggregation function $\phi^{a}$ must satisfy permutation invariance, 
i.e., a permutation transformation on the input $\bm{f}_{\mathcal{N}(i)}$ 
will have no influence on the output $f_i^{(1)}$. 

\textbf{Point update.} 
The point update usually contains an MLP, 
which is used to further enhance the point features. 

\textbf{Position embedding.} 
Incorporating representations of position information via position embedding 
is a common practice for Vision Transformer (ViT) \cite{rethinkrpe, deit, swin}, 
since ViT models must capture spatial relationships of tokens. 
We thus introduce the concept of position embedding into PointMeta. 
Within a point cloud, points are organized in an unstructured format, 
however the points are highly correlated in the 3D Euclidean space. 
Therefore, the building block of point cloud analysis should posses the position awareness, 
which is provided by the position embedding function $\phi^{e}$. 
The function $\phi^{e}$ is usually combined with 
the neighbor update function $\phi^{n}$ or the aggregation function $A$
implicitly or explicitly, thus we present them 
with an bundled notation, $\phi^{n}\circ \phi^{e}$ or $\phi^{a}\circ \phi^{e}$. 

One may associate PointMeta with 
Message Passing Neural Networks (MPNNs) \cite{MPNN},
which abstracts the calculations on a graph as a message passing process.  
However, it should be noted that PointMeta has a deeper and more exact abstraction 
on the neighbor calculations from two aspects 
compared with applying MPNN directly.   
First, PointMeta decouples the calculations on neighbor features 
into a neighbor update and a neighbor aggregation. 
Second, PointMeta emphasizes the importance of position awareness in the neighbor computation. 
In contrast, a single message function in MPNN is too rough 
to describe the calculations on neighbor points. 
As we will see in the later section, it is easy to use PointMeta 
to describe and compare the calculation process of the building block of existing models 
completely and thoroughly,  
but it is difficult for MPNN. 

\subsection{Instantiation Examples}
\label{instantiation}
In this section, we select three representative models from the literature 
as the instantiation examples and explain 
how these models implement the meta functions within PointMeta. 
We illustrate more instantiation examples \cite{dgcnn,pointcnn,randla,pointconv,kpconv,pointnext} in the supplementary due to the limited space. 
Sometimes these meta functions are implemented sophisticatedly, 
we will decompose them into several subfunctions for the sake of brevity and readability. 

\textbf{PointNet++}\cite{pointnet2}. 
The neighbor update is described as follows:
\begin{gather*}
\bm{f}_{\mathcal{N}(i)}^{\prime}, \bm{p}_{\mathcal{N}(i)} = \text{Group}(f_i, p_i),\\
f_j = \text{MLP}_1(\text{Concat}(f_j^{\prime}, p_j-p_i)), 
\end{gather*}
where $\text{MLP}_1$ denotes the multi-layer perceptions applied 
on the concatenated neighbor features. 
The neighbor aggregation is implemented using a simple max pooling operation,
$$
f_i^{(1)} = \text{Max}(\bm{f}_{\mathcal{N}(i)}). 
$$
The point update function is another MLP layer, thus we have
$$
f_i^{(2)} = \text{MLP}_2(f_i)
$$
Note that the position embedding function $\phi_e$ 
is implicitly implemented by $\text{MLP}_1$, 
which incorporates the concatenation of the relative positions $p_j-p_i$ 
and grouping features $f_j^{\prime}$ as input
and returns position-aware neighbor features $f_j$.

\textbf{Point Transformer} \cite{pointtransformer}.
The neighbor update is described as follows:
\begin{gather*}
 f_i^{\prime} = \text{MLP}_1(f_i),\\
 \bm{f}_{\mathcal{N}(i)}, \bm{p}_{\mathcal{N}(i)} = \text{Group}(f_i, p_i).
\end{gather*}
The neighbor aggregation is implemented using vector self-attention, 
which is described as
\begin{gather*}
e^j = \text{MLP}_2(p_j-p_i),\\
M_{j} = \text{Softmax}{(\text{MLP}_5 (\text{MLP}_3(f_i)-\text{MLP}_4(f_j) + e_j))},\\
f_i^{(1)} = \sum_{j\in \mathcal{N}(i)}(M_j\odot (f_j + e_j)),
\end{gather*}
where the $\odot$ denotes the Hadamard product 
and $M\in \mathbb{R}^{k\times d}$ denotes the vector self-attention (VSA) mask.  
Finally, the point feature is updated with an MLP layer and a shortcut, then we have
$$
f_i^{(2)} = \text{MLP}_6(f_i^{(1)}) + f_i
$$

\textbf{ASSANet}\cite{assanet}.
The neighbor update is described as:
\begin{gather*}
f_i^{\prime} = \text{MLP}_1(f_i),\\
\bm{f}_{\mathcal{N}(i)}^{\prime}, \bm{p}_{\mathcal{N}(i)} = \text{Group}(f_i, p_i),\\
f_j = \text{Repeat}_1(f_j^{\prime}),
\end{gather*}
where the operation $\text{Repeat}_1$ means repeating the features three times
to produce $f_j \in \mathbb{R}^{3d\times 1} $. 
The neighbor aggregation is implemented using an anistropic function, 
which is described as: 
\begin{gather*}
e^j = \text{Repeat}_2(p_j-p_i),\\
f_i^{(1)}=\frac{1}{k}\sum_{j\in \mathcal{N}(i)}(f_j \odot e_j), 
\end{gather*}
where the position embedding $e^j \in \mathbb{R}^{3d\times 1}$ is produced 
by simply repeating the relative coordinates $d$ times. 
Note that, a similar position embedding and aggregation method 
are also adopted by PosPool \cite{pospool}, which is named position pooling. 
Similar with Point Transformer \cite{pointtransformer}, 
the point feature is updated with an MLP layer and a shortcut.

\section{Explore Best Practices} 
Given how many instantiations of PointMeta have appeared in the literature, 
we should push this general framework as far as possible in the architecture exploration 
to determine the most crucial implementation details. 
With the help of PointMeta, it is convenient to use quick experiments with variables controlled to verify any empirical observations or assumptions summarized from the comparison. 
In this section, we first compare and analyze the building blocks of existing  models 
from the implementation details of these meta functions (Sec. \ref{sec_nu}, \ref{sec_pe}, \ref{sec_na}, \ref{sec_pu}). 
Then, we summarize the best practices from the systematic analysis (Sec. \ref{sec_practice}). 
Finally, applying these practices, we do simple tweaks on the building blocks and propose a simple yet well-performing block, PointMetaBase (Sec. \ref{pointmetabase}). 

\subsection{Neighbor Update}\label{sec_nu}
According to the order of  $\text{Group}$ and $\text{MLP}$ operation, 
the neighbor update function is categorized into two classes: 
\begin{itemize}
  \item $\text{Group}$ before $\text{MLP}$ \cite{pointnet2, dgcnn, deepgcn, pointnext}. 
  \item $\text{MLP}$ before $\text{Group}$ \cite{pointcnn, pospool, assanet, pointtransformer, stratified}. 
\end{itemize}
As pointed out by ASSANet \cite{assanet}, 
compared with the first order, 
adopting the second order reduces the FLOPs by 
$\frac{d \times d \times N \times K \times L}{d \times d \times N \times L} = K$ times, 
where $N$ denotes the point number, $K$ denotes the neighbor number, $L$ denotes the layer number of the MLP. 
Although applying MLP before Group brings a large efficiency gain for the neighbor update, 
a side effect is that the relative coordinates cannot be used as input for the MLP. 
However, this problem can be easily solved by other position embedding methods, 
which is specified in Sec. \ref{sec_pe}. 

\begin{table}[h]
  \centering
  \begin{tabular}{l|lll}
  \hline
  Variant       & \begin{tabular}[c]{@{}l@{}}mIoU\\ (\%)\end{tabular} & \begin{tabular}[c]{@{}l@{}}Params\\ (M)\end{tabular} & \begin{tabular}[c]{@{}l@{}}FLOPs\\ (G)\end{tabular} \\ \hline
  Plain-Max     & 47.3±0.7                                                & 2.0                                                  & 1.4                                                 \\ \hline
  Plain-PP      & 65.1±0.2                                                & 2.0                                                  & 1.5                                                 \\
  Plain-PP-Max  & 58.4±0.6                                                & 2.0                                                  & 1.5                                                 \\
  Plain-IPE-Max & 68.0±0.3                                                & 2.0                                                  & 13.8                                                \\
  Plain-EPE-Max & \textbf{69.0±0.3}                                       & 2.0                                                  & 1.8                                                 \\ 
  Plain-EPE-PP & 65.4±0.1                                                & 2.0                                                  & 1.8                                                 \\ \hline
  \end{tabular}
  \caption{Performance of the blocks with different position embedding functions 
  and neighbor aggregation function on Area 5 of S3DIS \cite{s3dis}. Plain-PP-Max means multiplying the neighbor features with relative coordinates as PP but aggregating via max pooling. }
  \label{tab_pe}
\end{table}

\subsection{Position Embedding}\label{sec_pe}
The manners of implementing the position embedding can be categorized into three classes: 
\begin{itemize}
  \item Implicit position embedding (IPE) \cite{pointnet2, dgcnn, deepgcn, pointnext}. 
  \item Explicit position embedding (EPE) \cite{pointcnn, pointtransformer, stratified}. 
  \item Position pooling (PP) \cite{pospool, assanet}.
\end{itemize}

\textbf{Implicit position embedding (IPE).} 
Following PointNet++ \cite{pointnet2}, 
many methods \cite{pointnext, pointmlp, pointweb} concatenate the relative coordinates 
with the neighbor features 
as inputs of the MLP layer in the neighbor update function. 
As going deeper through the networks, the dimension  of the neighbor features become higher 
while the coordinates dimension still remains 3. 
This may cause the loss of position information in the high-level features. 
In addition, IPE requires applying MLP before Group in the neighbor update function, 
which will increase the computation $K$ times.

\textbf{Explicit position embedding (EPE).} 
Incorporating explicit representations of position information 
is a common practice for ViT \cite{rethinkrpe, deit, swin}. 
As mentioned in Sec. \ref{instantiation}, Point Transformer \cite{pointtransformer} 
learn a position embedding $e^j$ using $\text{MLP}_2$ and add it to the neighbor features $f_j$.  
Compare with IPE, EPE avoid the dimension imbalance 
between the position representations and the neighbor features. 
Because the input dimension of $\text{MLP}_2$ is only 3, 
the computation of EPE become very small by controling the dimension of the hidden layers. 
PointCNN \cite{pointcnn} also uses an MLP to implement the EPE function, 
but it merges the EPE by concatenation instead of addition, 
which incurs a computation increase. 
StratifiedFormer \cite{stratified} maintains a learnable quantized look-up table as EPE.
In this section, we select the EPE of Point Transfomer 
as the representative EPE method for the later experiments 
due to its simplicity. 

\textbf{Position pooling (PP). } 
As mentioned in Sec. \ref{instantiation}, ASSANet \cite{assanet} and PosPool \cite{pospool}
use relative coordinates as weights to pool neighbor features. 
This way can embed the position information into the neighbor features 
and make an aggregation function without learnable weights. 
The efficiency of position pooling is worth affirming, 
but embedding the position in a non-data-driven way 
may harm the learned neighbor features. 

\textbf{Empirical analysis.} 
We conduct empirical experiments to evaluate the different PE functions described above. 
The first step is to construct a plain block as the baseline. 
Following the suggestion given in Sec. \ref{sec_nu}, 
the neighbor update function adopts the order of MLP-before-Group. 
The layer number of the MLP here is set to 1. 
The neighbor aggregation function is a simple max pooling. 
The point update function is a 1-layer MLP.    
We denote this block as Plain-Max. 
We equip Plain-Max with the three kinds of position embedding described above 
and compare their performance on Area 5 of S3DIS \cite{s3dis}. 
The MLP used respectively in EPE and IPE are both 1-layer. 
The macro-architecture is in level "L" (described in Sec. \ref{pointmetabase}). 
As shown in Tab. \ref{tab_pe}, Plain-EPE-Max achieves the best performance 
with a tiny computation increase against Plain-PP, Plain-PP-Max and Plain-IPE-Max. 
Because coupling the neighbor features transformation with position embedding, 
Plain-IPE incurs heavy redundant computation. 
In a summary, EPE is the optimal choice among the three kinds of position embeddings. 

\begin{table}[]
  \begin{tabular}{l|lll}
  \hline
  Variants                             & \begin{tabular}[c]{@{}l@{}}mIoU\\ (\%)\end{tabular} & \begin{tabular}[c]{@{}l@{}}Params\\ (M)\end{tabular} & \begin{tabular}[c]{@{}l@{}}FLOPs\\ (G)\end{tabular} \\ \hline
  Point Trans \cite{pointtransformer}  & 70.5±0.3                                            & 7.8                                                  & 5.6                                                 \\
  Point Trans (-VSA+Max)                  & 70.3±0.2                                            & 5.1                                                  & 3.3                                                 \\ \hline
  \end{tabular}
  \caption{Ablation of the vector self-attention of Point Transfomer \cite{pointtransformer} 
  on Area 5 of S3DIS \cite{s3dis}. -VSA+Max means replacing VSA with max pooling.}
  \label{tab_pt_ablated}
\end{table}

\subsection{Neighbor Aggregation}\label{sec_na}
According to whether learnable weights exist in the neighbor aggregation function, 
we divide the aggregations function into two types:
\begin{itemize}
  \item Non-learnable aggregation \cite{pointnet2, dgcnn, deepgcn, pospool, assanet, pointnext}. 
  \item Learnable aggregation \cite{pointcnn,pointtransformer,stratified,kpconv,pointconv}. 
\end{itemize}

\textbf{Learnable aggregation. } 
As mentioned in Sec. \ref{sec_meta_arch}, 
the aggregation function should satisfy permutation invariance because the point cloud is unordered. 
The permutation invariance is acquired by learning a space transformation \cite{pointcnn, pointconv}, 
learning pointwise weights \cite{pointtransformer}, or pairwise self-attention \cite{stratified}. 
In this section, we select the vector self-attention of Point Transfomer \cite{pointtransformer} 
as the representative implementation of the learnable aggregation function 
due to its excellent performance. 

\textbf{Non-learnable aggregation. }
There are two types of non-learnable aggregation in the literatures: 
position pooling \cite{pospool, assanet} 
and max pooling \cite{pointnet2, dgcnn, deepgcn, pointnext}. 
As mentioned in Sec. \ref{sec_pe}, position pooling embed the position in a non-data-driven way and may compete with the learned neighbor features. 
As shown in Tab. \ref{tab_pe}, Plain-EPE-Max surpasses Plain-EPE-PP significantly, 
which supports our assumption. 
Therefore, we select the max pooling 
as the representive non-learnable aggregation function. 

\textbf{Empirical analysis. }
We conduct an ablation experiment on Point Transformer \cite{pointtransformer} 
to compare the aggregation power of both max pooling and vector self-attention. 
Specially, we replace the VSA with a max pooling operation (denoted as -VSA+Max). 
As a result, the aggregation of the ablated Point Transfomer is formulated as: 
$$
e^j = \text{MLP}_2(p_j-p_i),
$$
\begin{equation}\label{eq_max_pt}
f_i^{(1)}=\text{Max}(\bm{f}_{\mathcal{N}(i)} + \bm{e}_{\mathcal{N}(i)}), j\in \mathcal{N}(i),  
\end{equation}
where $\bm{e}_{\mathcal{N}(i)}=\{e_j | j\in \mathcal{N}(i)\}$. 
As shown in Tab. \ref{tab_pt_ablated}, compared with VSA, 
max pooling achieves comparable performance with much fewer parameters and FLOPs. 
Here comes a question, why max pooling has a comparable power with the learnable aggregation? 
We explore the answer within our PointMeta framework and find that the max pooling 
is a special case of VSA but with binary and sparse attention. 
Specifically, the aggregation function Eq. \ref{eq_max_pt} can be reformulated as follows: 
\begin{gather*}
M_{j} = \text{Softmax}_{T\to 0}{(\bm{f}_{\mathcal{N}(i)})},\\
f_i^{(1)}=\sum_{j\in \mathcal{N}(i)}(M_{j}\odot (f_j + e_j)), 
\end{gather*}
where $\text{Softmax}_{T\to 0}$ denotes a softmax function 
whose temperature approximates 0, 
which makes the softmax output $M_{j}$ approximates the binary masks. 
This formulation reveals the collaboration between max pooling and the neighbor update function, 
i.e., max pooling is an aggregation function with binary and sparse attention which is generated by applying softmax on the output of the neighbor update

\subsection{Point Update}\label{sec_pu}
The point update function is used to enhance the point features after the neighbor aggregation. 
Several existing studies do not implement this function in their building blocks \cite{dgcnn, deepgcn}, 
and some other studies  use a heavy point update function, 
such as a 2-layer MLP with an inverted bottleneck \cite{stratified, pointnext}. 
A natural question arises: 
how much computation complexity should we allocate for the point update function? 
We answer this question via extensive ablated experiments. 

\textbf{Empirical analysis. }
According to the exploration above, we adopt the Plain-EPE-Max as the baseline 
due to its excellent performance in the ablated experiments (Tab. \ref{tab_pe}). 
Since the computation are concentrated on the MLP of 
the neighbor update function and point update function, 
we can allocate computation between the two update functions 
by changing the configurations of the two MLP. 
We ablate the MLP from two aspects, i.e., 
the layer number and whether to use an inverted bottleneck 
that adopted by PointNeXt \cite{pointnext} and StratifiedFormer \cite{stratified}. 
As shown in Tab. \ref{tab_mlp}, the configuration "N1P2" can achieve 
an optimal balance between complexity and performance. 
More layers or inverted bottlenecks can only promote the model marginally 
but lead to much more computation. 

\begin{table}[]
  \centering
  \begin{tabular}{l|lll}
    \hline
    Variants & \begin{tabular}[c]{@{}l@{}}mIoU\\ (\%)\end{tabular} & \begin{tabular}[c]{@{}l@{}}Params\\ (M)\end{tabular} & \begin{tabular}[c]{@{}l@{}}FLOPs\\ (G)\end{tabular} \\ \hline
    N1P1     & 69.0±0.3                                            & 2.0                                                  & 1.7                                                 \\
    N2P0     & 68.7±0.3                                            & 2.0                                                  & 1.7                                                 \\
    N0P2     & 67.8±0.4                                            & 2.0                                                  & 1.7                                                 \\
    N1P2     & 69.5±0.3                                            & 2.7                                                  & 2.0                                                 \\
    N1P2-Inv & \textbf{69.7±0.3}                                   & 7.1                                                  & 3.8                                                 \\
    N2P1     & 69.4±0.3                                            & 2.7                                                  & 2.0                                                 \\
    N2P1-Inv & 69.6±0.4                                            & 7.1                                                  & 3.8                                                 \\
    N1P3     & 69.5±0.3                                            & 3.4                                                  & 2.3                                                 \\
    N3P1     & 69.6±0.5                                            & 3.4                                                  & 2.3                                                 \\ \hline
    \end{tabular}
  \caption{Semantic segmentation of the Plain-EPE with different MLP configuration 
  on Area 5 of S3DIS \cite{s3dis}. 
  "N1P2" means that the MLP in the neighbor update is 1-layer and 2-layer. 
  "Inv" means the inverted bottleneck. }
  \label{tab_mlp}  
\end{table}

\subsection{Best Practices}\label{sec_practice}
Here we summarize the best practices for the building block as follows: 
\begin{itemize}
  \item \textbf{Postion Embedding}. 
  Explicit position embedding can endow the models with better position awareness with a low computation cost 
  comparing with other PE manners. 
  \item \textbf{Point Update}. 
  Point update is important and the complexity allocation between the neighbor update and the point update should be properly set. 
  \item \textbf{Neighbor Aggregation}. 
  Max pooling is not a pure non-learnable aggregation function, 
  its aggregation power is comparable with the learnable aggregation function 
  but with much less computation complexity. 
  Any exploration for a new aggregation function should set max pooling as the baseline. 
  \item \textbf{Neighbor Update}. 
  Applying the MLP before neighbors grouping will bring a $K$ times FLOPs reduction 
  comparing with using the contrary order. However, the position embeding  function must be carefully chosen for the postition awareness. 
\end{itemize}

The first and the second practice are usually ignored by researchers 
but are vital for the effectiveness and efficiency of the building block, 
which is demonstrated in Sec. \ref{sec_pe} and Sec. \ref{sec_pu}.
The third shares a similar viewpoint with PosPool \cite{pospool} 
which claims that different local aggregation operators perform similarly.   
However, from the perspective of PointMeta, we provide a deeper insight that max pooling can be reformulated 
as a special case of vector self-attention \cite{pointtransformer}, which makes it have comparable aggregation power 
comparing with the learnable aggregation function.  
The fourth practice about neighbor update has been emphasized by ASSANet \cite{assanet}, we restate it here  
because it is important to notice its collabration with position embedding function.  
We believe that these practices can provide a guideline for the future research. 

\subsection{PointMetaBase}
\label{pointmetabase}
We apply the best practices summarized in Sec. \ref{sec_practice} 
to design the building block 
and implement the four meta functions as simply as possible. 
We name this block as PointMetaBase, 
which we hope to become a solid baseline for future research. 
Fig. \ref{fig_meta_arch} (g) shows the architecture of PointMetaBase. 
The layer number of the first MLP is set to 1 and that of the second MLP is set to 2. 
The neighbor aggregation function is a simple max pooling. 
We use the EPE adopted by Point Transformer \cite{pointtransformer} as the position embedding function. 
Applying EPE and the MLP-before-Group order, we tweak the set abstraction module \cite{pointnet2, pointnext} as the reduction block, termed PointMetaSA. 
We adopt the same scaling strategies and decoder with PointNeXt \cite{pointnext} to construct our PointMetaBase family: 
\begin{itemize}
  \item $\text{PointMetaBase-S}: C=32, B=0$ 
  \item $\text{PointMetaBase-L}: C=32, B=(2, 4, 2, 2)$ 
  \item $\text{PointMetaBase-XL}: C=64, B=(3, 6, 3, 3)$ 
  \item $\text{PointMetaBase-XXL}: C=64, B=(4, 8, 4, 4)$ 
\end{itemize}
$C$ denotes the channel size of the stem MLP 
and $B$ denotes the number of the PointMetaBase block in each stage. 
Note that when $B = 0$, only one PointMetaSA block 
but no PointMetaBase blocks are used at each stage. 
Due to the excellent efficiency of PointMetaBase, 
we do not need to construct the network at the level "B" as done in PointNeXt \cite{pointnext}. 
In contrast, we choose to scale the model to a larger level "XXL". 
The full details of network architecture are avaliable in the supplementary materials.

\begin{table*}[ht]
  \centering
  \begin{tabular}{l|ll|ll|ll|llc}
    \toprule
    \multirow{3}{*}{Method} & \multicolumn{2}{c|}{S3DIS 6-Fold}                                                                       & \multicolumn{2}{c|}{S3DIS Area-5}                                                                       & \multicolumn{2}{c|}{ScanNet mIoU}                                                                            & Params. & FLOPs & Throughput  \\ \cline{2-7} 
                            & \begin{tabular}[c]{@{}l@{}}mIoU\\ (\%)\end{tabular} & \begin{tabular}[c]{@{}l@{}}OA\\ (\%)\end{tabular} & \begin{tabular}[c]{@{}l@{}}mIoU\\ (\%)\end{tabular} & \begin{tabular}[c]{@{}l@{}}OA\\ (\%)\end{tabular} & \begin{tabular}[c]{@{}l@{}}Val\\ (\%)\end{tabular} & \begin{tabular}[c]{@{}l@{}}Test\\ (\%)\end{tabular} & M       & G     & (ins./sec.) \\ \hline
    PointNet++ \cite{pointnet2}               & 54.5                                                & 81.0                                              & 53.5                                                & 83.0                                              & 53.5                                                & 55.7                            & 1.0     & 7.2   & 237         \\
    PointCNN \cite{pointcnn}                  & 65.4                                                & 88.1                                              & 57.3                                                & 85.9                                              & -                                                   & 45.8                            & 0.6     & -     & -           \\
    DeepGCN \cite{deepgcn}                    & 60.0                                                & 85.9                                              & 52.5                                                & -                                                 & -                                                   & -                               & 3.6     & -     & -           \\
    KPConv \cite{kpconv}                      & 70.6                                                & -                                                 & 67.1                                                & -                                                 & 69.2                                                & 68.6                            & 15.0    & -     & -           \\
    RandLA-Net \cite{randla}                  & 70.0                                                & 88.0                                              & -                                                   & -                                                 & -                                                   & 64.5                            & 1.3     & 5.8   & -           \\
    BAAF-Net \cite{baaf}                      & 72.2                                                & 88.9                                              & 65.4                                                & 88.9                                              & -                                                   & -                               & 5.0     & -     & -           \\
    Point Transformer \cite{pointtransformer} & 73.5                                                & 90.2                                              & 70.4                                                & 90.8                                              & 70.6                                                & -                               & 7.8     & 5.6   & -           \\
    CBL \cite{cbl}                            & 73.1                                                & 89.6                                              & 69.4                                                & 90.6                                              & -                                                   & 70.5                            & 18.6    & -     & -           \\ 
    ASSANet \cite{assanet}                    & -                                                   & -                                                 & 65.8                                                & 88.9                                              & -                                                   & -                               & 2.4     & 2.5   & 300         \\
    ASSANet-L \cite{assanet}                  & -                                                   & -                                                 & 68.0                                                & 89.7                                              & -                                                   & -                               & 115.6   & 36.2  & 136         \\ \hline
    PointNeXt-S \cite{pointnext}              & 68.0                                                & 87.4                                              & 63.4±0.8                                            & 87.9±0.3                                          & 64.5                                                & -                               & 0.8     & 3.6   & 276         \\
    PointNeXt-B \cite{pointnext}              & 71.5                                                & 88.8                                              & 67.3±0.2                                            & 89.4±0.1                                          & 68.4                                                & -                               & 3.8     & 8.9   & 161         \\
    PointNeXt-L \cite{pointnext}              & 73.9                                                & 89.8                                              & 69.0±0.5                                            & 90.0±0.1                                          & 69.4                                                & -                               & 7.1     & 15.2  & 109         \\
    PointNeXt-XL \cite{pointnext}             & 74.9                                                & 90.3                                              & 70.5±0.3                                            & 90.6±0.1                                          & 71.5                                                & 71.2                            & 41.6    & 84.8  & 43          \\ \hline
 \rowcolor{gray!10} \textbf{PointMetaBase-L}           & \textbf{75.6}                       & \textbf{90.6}                                     & 69.5±0.3                                            & 90.5±0.1                                          & 71.0                                                & -                               & 2.7     & 2.0   & 187         \\
  \rowcolor{gray!10}  \textbf{PointMetaBase-XL}          & \textbf{76.3}                     & \textbf{91.0}                                     & \textbf{71.1±0.4}                                   & \textbf{90.9±0.1}                                 & 71.8                                                & -                               & 15.3    & 9.2   & 104         \\ 
  \rowcolor{gray!10}  \textbf{PointMetaBase-XXL}         & \textbf{77.0}                     & \textbf{91.3}                                     & \textbf{71.3±0.7}                                   & \textbf{90.8±0.6}                                 & 72.8                                                & 71.4                            & 19.7    & 11.0  & 90          \\ \bottomrule
    \end{tabular}
    \caption{Semantic segmentation on S3DIS \cite{s3dis} (6-Fold and Area 5) and ScanNet V2 \cite{scannet}. Our PointMetaBase family surpass the state-of-the-art method  PointNeXt \cite{pointnext} significantly with large FLOPs reduction. }
    \label{tab_s3dis}
\end{table*}

\section{Experiments}
\label{sec_exps}
We studied the performance of PointMetaBase 
on S3DIS \cite{s3dis}, ScanNet V2 \cite{scannet}, ScanObjectNN \cite{scanobjectnn} and ShapeNetPart \cite{shapenetpart}. 
To enable a fair comparison, 
the same data processing and evaluation protocols adopted 
by the state-of-the-art method PointNeXt \cite{pointnext} were used in our experiments.

\textbf{Experimental Setups.} 
We optimized all our models using CrossEntropy loss with label smoothing \cite{labelsmooth}, 
AdamW optimizer, an initial learning rate 0.001, 
and weight decay $10^{-4}$, with Cosine Decay, 
with a 32G V100 GPU. 
Model parameters, FLOPs and inference throughput 
are provided for comparison. 
For a fair comparison, we use the open codes
provided by PointNext \cite{pointnext} to conduct the measurements.
\footnote{
https://github.com/guochengqian/PointNeXt
}
We also use MindSpore to validate the generalization on different deep learning frameworks.
All the measurements are conducted 
using an NVIDIA Tesla A100 40GB GPU and a 12-core Intel Xeon @ 2.40GHz CPU.

\subsection{3D Scene Segmentation}
S3DIS \cite{s3dis} 
is a challenging benchmark composed of 
6 large-scale indoor areas, 271 rooms, and 13 semantic categories in total. 
The standard 6-fold cross-validation results and the Area-5 results on S3DIS are reported in Tab. \ref{tab_s3dis}.  
Since we adopt the same scaling strategies, 
PointMetaBase-L and PointMetaBase-XL have the same configuration on width and depth 
as PointNeXt-L and PointNeXt-XL. 
With only 13\% FLOPs, 
our PointMetaBase-L outperforms PointNext-L by 1.7\% and 0.8\% 
in terms of mean IoU (mIoU) and overall accuracy (OA), respectively,
and is much faster in terms of throughput. 
Compared  with PointNext-XL, 
our PointMetaBase-XL uses only 11\% of the FLOPs, but 
surpasses with 1.4\% mIoU and 0.7\% OA. 
The inference speed is $2.4\times$ faster. 
The excellent efficiency and performance of PointMetaBase can be attributed to the
the removal of the computation redundancy in the two update functions 
and the improved position awareness introduced by explicit position embedding. 
ScanNet \cite{scannet} contains 3D indoor scenes of various rooms with 20 semantic categories. The dataset is divided into 3 part: training, validation, and test splits, with 1201, 312 and 100 scans, respectively. Following PointNeXt \cite{pointnext}, we do not use any voting strategies. 
As shown in Tab. \ref{tab_s3dis}, our PointMetaBase-L surpasses PointNeXt-L 1.6\% mIoU with only 14\% FLOPs and 1.7$\times$ faster inference speed. PointMetaBase-XL surpasses PointNeXt-XL 0.3\% mIoU with only 11.3\% FLOPs and 2.4$\times$ faster inference speed. 
Our largest variant, PointMetaBase-XXL, achieves better performance.

\begin{table*}[]
  \centering
\begin{tabular}{l|ll|lll}
\hline
Method                                  & \begin{tabular}[c]{@{}l@{}}ins. mIoU\\ (\%)\end{tabular} & \begin{tabular}[c]{@{}l@{}}cls. mIoU\\ (\%)\end{tabular} & \begin{tabular}[c]{@{}l@{}}Params\\ (M)\end{tabular} & \begin{tabular}[c]{@{}l@{}}FLOPs\\ (G)\end{tabular} & \begin{tabular}[c]{@{}l@{}}Throughput\\ (ins./sec)\end{tabular} \\ \hline
PointNet++ \cite{pointnet2}                             & 85.1                                                    & 81.9                                                     & 1.0                                                  & 4.9                                                 & -                                                               \\
KPConv \cite{kpconv}                                 & 86.4                                                    & 85.1                                                     & -                                                    & -                                                   & -                                                               \\
ASSANet \cite{assanet}                                & 86.1                                                    & -                                                        & -                                                    & -                                                   & -                                                               \\
Point Transformer \cite{pointtransformer}                       & 86.6                                                    & 83.7                                                     & 7.8                                                  & -                                                   & -                                                               \\
PointMLP \cite{pointmlp}                               & 86.1                                                    & 84.6                                                     & -                                                    & -                                                   & -                                                               \\
StratifiedFormer \cite{stratified}                       & 86.6                                                    & 85.1                                                     & -                                                    & -                                                   & -                                                               \\ \hline
PointNeXt-S  \cite{pointnext}                           & 86.7±0.0                                                & 84.4±0.2                                                 & 1.0                                                  & 4.5                                                 & 873                                                             \\
PointNeXt-S (C=64)  \cite{pointnext}                    & 86.9±0.1                                                & 84.8±0.5                                                 & 3.7                                                  & 17.8                                                & 391                                                             \\
PointNeXt-S (C=160)  \cite{pointnext}                   & 87.0±0.1                                                & \textbf{85.2±0.1}                                                 & 22.5                                                 & 110.2                                               & 118                                                             \\ \hline
\textbf{PointMetaBase-S (ours)}                  & 86.7±0.0                                                & 84.3±0.1                                                 & 1.0                                                  & 1.39                                                & 1194                                                            \\
\textbf{PointMetaBase-S (C=64) (ours)}  & 86.9±0.1                                                & 84.9±0.2                                                 & 3.8                                                  & 3.85                                                & 706                                                             \\
\textbf{PointMetaBase-S (C=160) (ours)} & \textbf{87.1±0.0}                                       & 85.1±0.3                                        & 22.7                                                 & 18.45                                               & 271                                                             \\ \hline
\end{tabular}
\caption{Part segmentation on ShapeNetPart \cite{shapenetpart}. Compaerd with the state-of-the-art method  PointNeXt \cite{pointnext}, our PointMetaBase family achieve comparable performance with large FLOPs reduction and faster inference speed. }
\label{tab_shapenetpart}
\end{table*}

\begin{table}[]
  \centering
  \begin{tabular}{l|lllll}
  \toprule
  Variants                        & \begin{tabular}[c]{@{}l@{}}OA\\ (\%)\end{tabular}  & \begin{tabular}[c]{@{}l@{}}FLOPs\\ (G)\end{tabular} & \begin{tabular}[c]{@{}l@{}}TP\\ (ins./sec)\end{tabular} \\ \hline
  PointNet++ \cite{pointnet2}     & 77.9                                              & 1.7                                                 & 2655                                                            \\
  PointCNN \cite{pointcnn}        & 78.5                                              & -                                                   & -                                                               \\
  DGCNN \cite{dgcnn}              & 78.1                                              & 4.8                                                 & 735                                                             \\
  DRNet \cite{pointnet}           & 80.3                                              & -                                                   & -                                                               \\
  PointMLP \cite{pointmlp}        & 85.4±1.3                                          & 31.3                                                & 405                                                             \\ \hline
  PointNeXt-S \cite{pointnext}    & 87.7±0.4                                          & 1.7                                                 & 2230                                                            \\
  \rowcolor{gray!10} \textbf{PointMetaBase-S} & \textbf{87.9±0.2}                     & 0.6                                                 & 2674                                                            \\ \bottomrule
  \end{tabular}
  \caption{Classification on ScanObjectNN \cite{scanobjectnn}. 
  Our model surpasses PointNeXt-S \cite{pointnext} both with much less computation cost. 
  }
  \label{tab_scan}
  \end{table}

\subsection{3D Object Classification}
ScanObjectNN \cite{scanobjectnn} contains about 15000 real scanning objects, 
which are grouped into 15 classes and have 2902 unique instances. 
Following \cite{pointmlp, pointnext}, 
we experiment on the hardest perturbed variant (PB\_T50\_RS). 
To make a fair comparison  with PointNeXt \cite{pointnext}, 
we did not  use upscaled variants of PointMetaBase on this benchmark.  
As reported in Tab. \ref{tab_scan}, 
our model surpasses PointNeXt-S, 
while using much fewer model FLOPs and running much faster. 

\subsection{3D Part Segmentation}
ShapeNetPart \cite{shapenetpart} is a dataset for part segmentation of objects.   
It consists of 16880 models with 16 different shape categories. 
Each category has 2-6 parts and the number of total part labels is up to 50. 
Following the state-of-the-art method PointNeXt \cite{pointnext}, we used voting by averaging the results of 10 randomly scaled input point clouds, with scaling factors equal to [0.8,1.2].
As shown in Tab. \ref{tab_shapenetpart}, compaerd with PointNeXt, our PointMetaBase family achieve comparable performance with large FLOPs reduction and faster inference speed. 
Specially, for C=32/C=64/C=160, the FLOPs of our PointMetaBase are only 30\%21/\%/17\% that of PointNeXt, and the inference speed of PointMetaBase is 1.4/1.8/2.3 times faster than that of PointNeXt. 

\section{Conclusion}
\label{sec_conclusion}
In this paper, we reformulate existing models for point cloud analysis 
into a general framework, PointMeta. 
Within this framework, 
we provide an in-depth analysis on the building blocks of existing models 
and summarize a few best practices for the building block design. 
Furthermore, we follow these best practices and propose a simple building block, PointMetaBase, 
which enjoys excellent performance and efficiency. 
The proposed framework and the simple block 
can encourage further rethinking and understanding on the building block design 
and shed new light on future architecture exploration. 


{\small
\bibliographystyle{ieee_fullname}
\bibliography{egbib}
}

\clearpage

\section*{Supplementary Material}

\section{Limitation and Future Work}
We further discuss the limitations of our work, which will be the focus in future. 
Firstly, although we find that explicit position embedding will promote performance significantly compared with other manners of position embedding, the rationale still remains unexplored. 
One possible direction is to look for theoretical explanations from gradient flows. 
Secondly, for the neighbor aggregation functions, we categorize them into learnable and non-learnable aggregation and find that the max pooling is a strong parameter-free method that is comparable to the learnable aggregation. 
Therefore, the next step is to do a finer taxonomy and richer ablation studies to find the applicable aggregation functions in different scenarios.

\section{More Instantiation Examples}
\textbf{DGCNN}\cite{dgcnn}.
The neighbor update is described as:
\begin{gather*}
\bm{f}_{\mathcal{N}(i)}^{\prime} = \text{Group}(f_i),\\
f_j = \text{MLP}_1(\text{Concat}(f_j^{\prime}-f_i, f_i)).
\end{gather*}
Different from PointNet++ \cite{pointnet2} that groups the neighbors in coordinate space, 
the neighbors grouping in DGCNN is conducted in the feature spaces. 
Similar to PointNet++, 
DGCNN uses max pooling as the neighbor aggregation function.  
The point update is absent in the DGCNN block. 
Additionally, there exists no position embedding function $\phi_e$ 
neither explicitly nor implicitly. 
In another word, there is no relative position information encoded in the output features. 

\textbf{PointCNN}\cite{pointcnn}. 
The neighbor update is described as:
\begin{gather*}
\bm{f}_{\mathcal{N}(i)}^{\prime}, \bm{p}_{\mathcal{N}(i)} = \text{Group}(f_i, p_i),\\
e_j = \text{MLP}_1(p_j-p_i),\\
f_j = \text{Concat}(f_j^{\prime}, e_j).
\end{gather*}
The neighbor aggregation is defined as: 
\begin{gather*}
\mathcal{X} = \text{MLP}_2(\bm{p}_{\mathcal{N}(i)}-p_i),\\
f_i^{(1)}=\text{Conv}(\bm{K}, \mathcal{X}\times \bm{f}_{\mathcal{N}(i)}), 
\end{gather*}
where $\times$ denotes matrix multiplication and 
$\bm{K}$ denotes the trainable convolution kernels of the convolution layer $\text{Conv}$.
Two position embedding functions exist in this block.  
The first one $\phi_1^e$ is $\text{MLP}_1$ used in the neighbor update, 
while the second one $\phi_2^e$ is implicitly implemented 
by $\text{Conv}(\bm{K}, \cdot)$ used in the neighbor aggregation function. 
Implementing position embedding by convolution layers
is a common practice in the ViT family \cite{cpvt,segformer}. 

\textbf{RandLA-Net} \cite{randla}.
The neighbor update is described as follows:
\begin{gather*}
\bm{f}_{\mathcal{N}(i)}^{\prime}, \bm{p}_{\mathcal{N}(i)} = \text{Group}(f_i, p_i),\\
f_j = \text{MLP}_1(\text{Concat}(f_j^{\prime},p_j)).
\end{gather*}
The position embedding is described as: 
$$
e^j = \text{MLP}_2(\text{Concat}(p_j-p_i,|p_j-p_i|, p_j, p_i)).
$$
The neighbor aggregation is implemented using attentive pooling that is similar to Point Transformer \cite{pointtransformer}.  
The neighbor aggregation is as follows:
\begin{gather*}
M_j = \text{Softmax}{(\text{MLP}_3(\text{Concat}(f_j, e_j)))},\\
f_i^{(1)}=\sum_{j\in \mathcal{N}(i)}(M_j \odot \text{Concat}(f_j, e_j)), 
\end{gather*}
where $M_j\in \mathbb{R}^{2d}$ denotes the attention weights for the neighbor $j$, 
$\odot$ denotes the Hadamard product.  
Finally, the point feature is updated with an MLP layer and a shortcut, then we have
$$
f_i^{(2)} = \text{MLP}_4(f_i^{(1)}) + f_i.
$$

\textbf{PointConv} \cite{pointconv}.
The neighbor update and point update of PointConv is same with PointNet++ \cite{pointnet2}. 
The original neighbor aggregation function of PointConv is defined as follow:
\begin{gather*}
g_j = \text{MLP}_1(p_j-p_i),\\
f_i=\sum_{j\in \mathcal{N}(i)}g_j \times f_j,
\end{gather*}
where $g_j \in \mathbb{R}^{d_{out}\times d_{in}}$ is the weight matrices that map features from dimension $d_{in}$ to $d_{out}$, $\times$ denotes matrix multiplication, L denotes the number of the kernel points,  $\{p_l|l<L \}$ denotes the coordinates of the kernel points, and $\{W_l|l<L \}$ denotes the associated weights matrices.
However, different with the convolution for 2D images, the weight matrices in each local neighborhood for 3D is unique, which results in a huge memory cost. To address this challenge, the authors propose an efficient version, which decouples the weight matrix for each local neighborhood into two parts: a dynamic weight matrix $M_j\in \mathbb{R}^{d_{mid}}$ and a static weight matrix $H\in \mathbb{R}^{d_{out}\times (d_{in}\times d_{mid})}$. 
In this way, the memory consumption of the generated weights reduces to $\frac{d_{mid}}{Nd_{out}}$ of the original version. 
The efficient version of neighbor aggregation is described as follows: 
$$
{M_j} = \text{MLP}_1(p_j-p_i),
$$
\begin{equation}\label{equ_pointconv1}
f_i^{\prime}=\sum_{j\in \mathcal{N}(i)}M_j \times f_j^{\top    },
\end{equation}
\begin{equation}\label{equ_pointconv2}
f_i = H\times \text{Vec}(f_i^{\prime}). 
\end{equation}
In Eq. \ref{equ_pointconv1}, the neighbor features $\{f_j|j\in \mathcal{N}(i)\}$ is transfomed into $f_i^{\prime}\in \mathbb{R}^{d_{mid}\times d_{in}}$. 
Then, in Eq. \ref{equ_pointconv2}, $f_i^{\prime}$ is turned into a vector by $\text{Vec}(\cdot)$ and then multiply with matrix $H$. 

\textbf{KPConv} \cite{kpconv}.
The neighbor update is described as follows:
$$
f_j, p_j = \text{Group}(f_i, p_i).
$$
The neighbor aggregation is as follows:
\begin{gather*}
g_j = \sum_{l=1}^{L}\max(0, 1-\frac{||p_j-p_l||}{\sigma})W_l,
f_i=\sum_{j\in \mathcal{N}(i)}g_j \times f_j,
\end{gather*}

where $\times$ denotes matrix multiplication, $L$ denotes the number of the kernel points,  $\{p_l|l<L \}$ denotes the coordinates of the kernel points, and $\{W_l|l<L \}$ denotes the associated weights matrices. 
Similar with the original version of PointConv \cite{pointconv}, the dynamic weights matrices $\{g_j|j\in \mathcal{N}(i)\}$ across all local neighborhoods will incur huge memory comsumption. 
Thus in the code implementation KPConv adopt similar strategy that first transforms the neighbor features with dynamic weights and then update point feature with static weights.  

\textbf{PointNeXt}\cite{pointnext}.
PointNeXt conducts almost the same neighbor update with PointNet++ \cite{pointnet2} 
except for the configuration of the $\text{MLP}$. 
PointNet++ applies a 3-layer MLP while PointNeXt only applies 1 layer 
on the neighbor feature $f_j^{\prime}$ to reduce computation. 
Besides, PointNeXt also adopts a simple max pooling operation. 
As for the point update, PointNeXt uses a 2-layer MLP with an inverted bottleneck 
together with a shortcut layer from the input point feature $f_i$, then we have:
$$
f_i^{(2)} = \text{MLP}_{inv}(f_i^{(1)}) + f_i.
$$

\begin{figure*}[ht!]
  \centering 
  \includegraphics[width=1.0\linewidth]{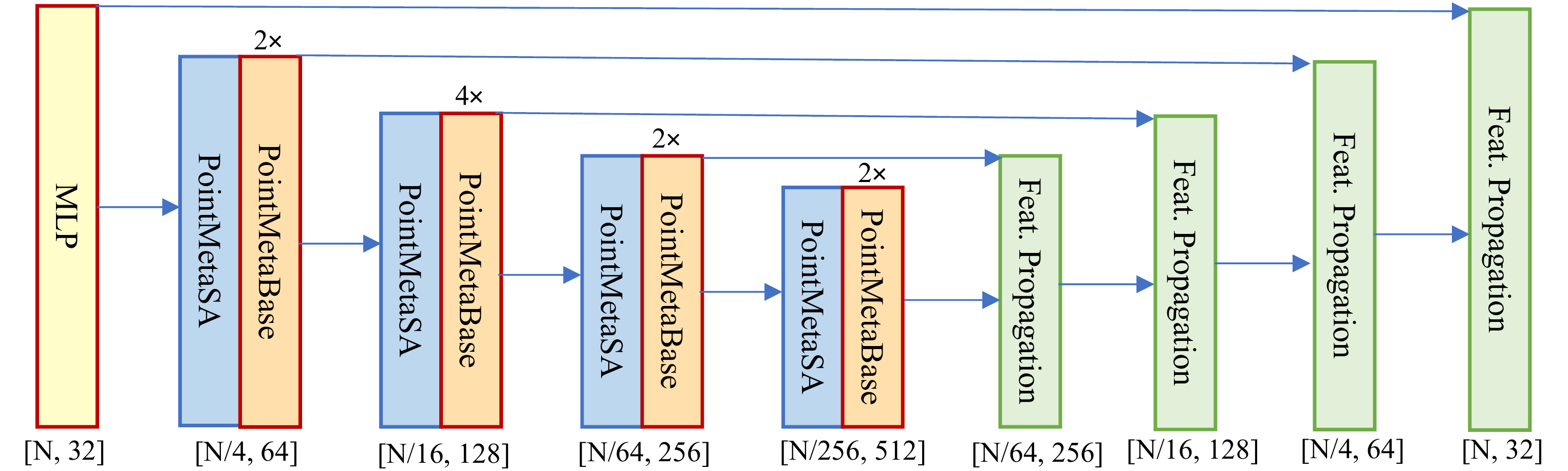}
  \hspace{-1.0em}
  \vspace{-0.6em}
  \caption{Macro-architecture of PointMetaBase-L. 
    Applying explicit position embedding and the MLP-before-Group order, we tweak the set abstraction module \cite{pointnet2,pointnext} as the reduction block, termed PointMetaSA. 
    We adopt the same scaling strategies and decoder with PointNeXt \cite{pointnext} to construct our PointMetaBase family.}
  \label{fig_macro_arch} 
\end{figure*}

\begin{figure*}[t]
  \centering 
  \includegraphics[width=1.0\linewidth]{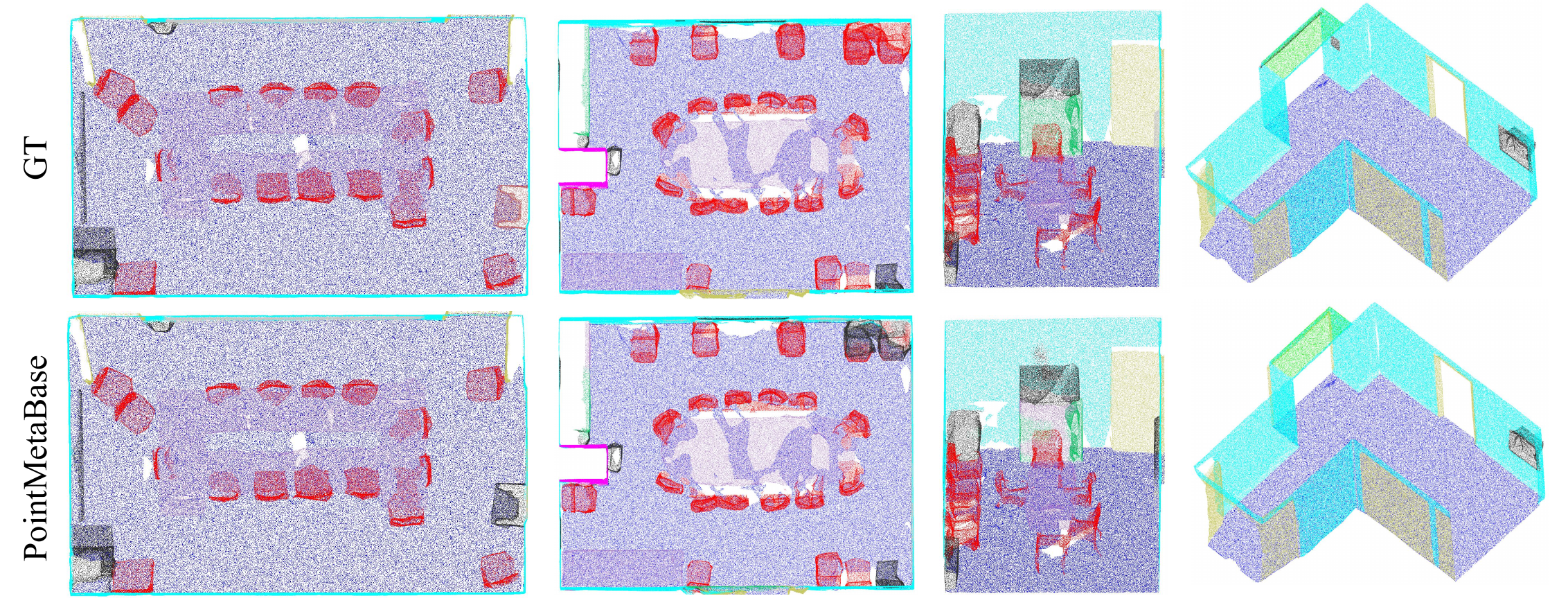}
  \hspace{-1.0em}
  \vspace{-0.6em}
  \caption{Semantic segmentation results on S3DIS \cite{s3dis}. The fist row is groundtruth, and the second one is predicted by PointMetaBase-XL. Best viewed in color.}
  \label{fig_vis} 
\end{figure*}

\section{Macro-Architecture}
Applying explicit position embedding and the MLP-before-Group order, we tweak the set abstraction module \cite{pointnet2,pointnext} as the reduction block, termed PointMetaSA. 
The decoder is composed of a series of feature propagation blocks \cite{pointnet2,pointnext}. 
As shown in Fig. \ref{fig_macro_arch}, 
a 1-layer MLP is used as stem in the first stage. 
In the later stages, PointMetaSA is placed first and the following are several PointMetaBase blocks.  
We adopt the same scaling strategies with PointNeXt \cite{pointnext} to construct our PointMetaBase family.
The configuration of our PointMetaBase family is summarized as follows: 
\begin{itemize}
  \item $\text{PointMetaBase-S}: C=32, B=0$ 
  \item $\text{PointMetaBase-L}: C=32, B=(2, 4, 2, 2)$ 
  \item $\text{PointMetaBase-XL}: C=32, B=(3, 6, 3, 3)$ 
  \item $\text{PointMetaBase-XXL}: C=32, B=(4, 8, 4, 4)$ 
\end{itemize}
$C$ denotes the channel size of the stem MLP 
and $B$ denotes the number of the PointMetaBase block in each stage. 
Note that when $B = 0$, only one PointMetaSA block 
but no PointMetaBase blocks are used at each stage. 
Due to the excellent efficiency of PointMetaBase, 
we do not need to construct the network at the level "B" as done in PointNeXt \cite{pointnext}. 
In contrast, we choose to scale the model to a larger level "XXL".


\end{document}